\documentclass[conference]{IEEEtran}
\IEEEoverridecommandlockouts

\usepackage{cite}
\usepackage{amsmath,amssymb,amsfonts}
\usepackage{algorithmic}
\usepackage{graphicx}
\usepackage{textcomp}
\usepackage{textcomp}
\usepackage{csquotes}
\usepackage{xcolor,soul}
\usepackage{adjustbox}
\usepackage{hyperref}
\usepackage{multirow}
\usepackage{titlesec}
\usepackage{subcaption}
\usepackage{makecell}
\usepackage{orcidlink}
\usepackage{booktabs} 
\usepackage{caption} 
\def\BibTeX{{\rm B\kern-.05em{\sc i\kern-.025em b}\kern-.08em
    T\kern-.1667em\lower.7ex\hbox{E}\kern-.125emX}}

\DeclareMathOperator{\argmin}{argmin}
\begin{document}

\title{Gaussian Process Regression for Improved Underwater Navigation
}

\author{\IEEEauthorblockN{1\textsuperscript{st} Nadav~Cohen \orcidlink{0000-0002-8249-0239}}
\IEEEauthorblockA{\textit{Hatter Department of Marine Technologies,} \\
\textit{Charney School of Marine Sciences,}\\
\textit{University of Haifa,}\\
Haifa, Israel \\
ncohe140@campus.haifa.ac.il}
\and
\IEEEauthorblockN{2\textsuperscript{nd} Itzik Klein \orcidlink{0000-0001-7846-0654}}
\IEEEauthorblockA{\textit{Hatter Department of Marine Technologies,} \\
\textit{Charney School of Marine Sciences,}\\
\textit{University of Haifa,}\\
Haifa, Israel \\
kitzik@univ.haifa.ac.il }

}

\maketitle

\begin{abstract}
Accurate underwater navigation is a challenging task due to the absence of global navigation satellite system signals and the reliance on inertial navigation systems that suffer from drift over time. Doppler velocity logs (DVLs) are typically used to mitigate this drift through velocity measurements, which are commonly estimated using a parameter estimation approach such as least squares (LS). However, LS works under the assumption of ideal conditions and does not account for sensor biases, leading to suboptimal performance. This paper proposes a data-driven alternative based on multi-output Gaussian process regression (MOGPR) to improve DVL velocity estimation. MOGPR provides velocity estimates and associated measurement covariances, enabling an adaptive integration within an error-state Extended Kalman Filter (EKF). We evaluate our proposed approach using real-world AUV data and compare it against LS and a state-of-the-art deep learning model, BeamsNet. Results demonstrate that MOGPR reduces velocity estimation errors by approximately 20\% while simultaneously enhancing overall navigation accuracy, particularly in the orientation states. Additionally, the incorporation of uncertainty estimates from MOGPR enables an adaptive EKF framework, improving navigation robustness in dynamic underwater environments. 
\end{abstract}


\section{Introduction}
\noindent
Underwater navigation is of great interest due to its potential applications and the challenges it presents. The underwater domain imposes significant navigation challenges, primarily because the global navigation satellite system (GNSS) signals cannot penetrate water and thus cannot be used for global localization. Additionally, large robotic platforms are typically employed due to extreme depths, high pressure, temperature variations, and low visibility. These platforms include autonomous underwater vehicles (AUVs), remotely operated vehicles (ROVs), and underwater gliders~\cite{sun2021review}.

\noindent
Underwater navigation systems commonly rely on an inertial navigation system (INS) and a Doppler velocity log (DVL). The INS estimates the navigation solution, namely the position, velocity, and orientation, in a global frame by integrating the inertial measurements through dead-reckoning. However, due to inherent stochastic errors and other error terms, this navigation solution accumulates errors over time. A DVL is typically used to mitigate this drift. The DVL transmits acoustic beams toward the seabed and utilizes the Doppler effect to estimate the vehicle’s velocity relative to the seabed. This velocity information is then integrated into a nonlinear filtering framework to correct the INS drift \cite{titterton2004strapdown}. The most common nonlinear filters used in INS/DVL fusion are the extended Kalman filter (EKF) \cite{fukuda2021performance} and the unscented Kalman filter (UKF) \cite{bucci2022evaluation}. 

\noindent
In a loosely coupled approach, the platform velocity vector from the DVL is obtained by applying a parameter estimation approach such as a least squares (LS) estimator to the raw beam measurements. Under Gaussian assumptions, LS is considered the best unbiased estimator, as it achieves the Cramér-Rao lower bound (CRLB). However, if the fundamental assumptions are not met—such as in the presence of biases, the LS solution is no longer optimal, leaving room for improvement \cite{bar2004estimation}.

\noindent
With recent advancements in hardware and the growing interest in data-driven and deep learning applications, machine learning methods have begun to be explored for inertial navigation \cite{cohen2024inertial}. Specifically, in the context of INS/DVL fusion, some data-driven approaches have started to emerge. Our previous work, BeamsNet, introduced an end-to-end, one-dimensional convolutional neural network model designed to replace the LS estimator for DVL velocity estimation \cite{cohen2022beamsnet}. In addition, BeamsNet was designed to handle scenarios where the DVL provides only partial measurements due to environmental conditions \cite{cohen2022libeamsnet,cohen2024Seamless}.
In \cite{zhang2020navnet}, the authors introduced NavNet, a neural network architecture that leverages a recurrent neural network (RNN) combined with an attention mechanism and fully connected (FC) layers to estimate AUV displacement. Various RNN-based models have been explored for underwater navigation, including hybrid gated RNNs, which eliminate the need for predefined motion models, thereby reducing modeling errors commonly encountered in traditional navigation algorithms. Other notable approaches include hybrid-RNN \cite{mu2019end}, long short-term memory (LSTM) networks \cite{topini2020lstm}, and position estimation using RNNs \cite{lv2021position}. Additionally, a radial basis function-based neural network has been proposed for INS/DVL fusion \cite{liu2022sins}. More recently, \cite{lv2024integration} introduced a deep sequence learning-based virtual GPS model integrated into an EKF framework to enhance AUV navigation accuracy.

\noindent
Another promising machine learning tool is Gaussian process regression (GPR), a probabilistic, non-parametric method for regression and function approximation. GPR provides predictions and estimates uncertainty, making it particularly advantageous for small datasets and applications requiring smooth and interpretable predictions without the need for a predefined model structure \cite{williams2006gaussian,murphy2023probabilistic}. GPR has been widely used in robotics for state estimation, sensor fusion, path planning, and learning-based control. Its ability to model uncertainties makes it highly valuable for autonomous systems, improving navigation accuracy, decision-making, and adaptability in real-world environments \cite{lilge2022continuum,west2021use,jakkala2024multi}. Recently, Chen \emph{et al.} utilized GPR to address the challenge of significant DVL outage gaps within INS/DVL sensor fusion \cite{chen2024seamless}.

\noindent
This paper proposes using multi-output GPR (MOGPR) to replace the LS estimator in DVL velocity estimation. As previous data-driven approaches have demonstrated, there is room for improvement since the LS estimator, in real-life scenarios,  is not truly unbiased and does not always meet the conditions for optimal estimation. The key advantage of MOGPR over alternative data-driven approaches is that it provides both velocity estimates (mean) and measurement covariance (uncertainty), which is essential for a reliable navigation system. This covariance can be incorporated into the sensor fusion filter, further enhancing the overall navigation solution.

\noindent
The main contributions of this paper are:
\begin{enumerate}
    \item A MOGPR approach designed explicitly for DVL measurements to enhance the platform velocity vector estimation.
    \item Integration of the MOGPR and the EKF in a loosely coupled manner, utilizing the MOGPR-estimated covariance within the filter for a fully adaptive and enhanced navigation solution.
\end{enumerate}

\noindent
The rest of the paper is organized as follows: Section \ref{sec:insdvl} provides an overview of INS/DVL sensor fusion, while section \ref{sec:MOGPR} introduces the MOGPR framework. The proposed approach is presented in section \ref{sec:sa}, and the results are summarized in Section \ref{sec:res}, where we discuss data acquisition and compare the performance of different approaches. Finally, conclusions are presented in section \ref{sec:con}.

\section{INS/DVL Fusion}\label{sec:insdvl}
\begin{figure}[htbp]
    \centering
    \includegraphics[width=0.8\columnwidth]{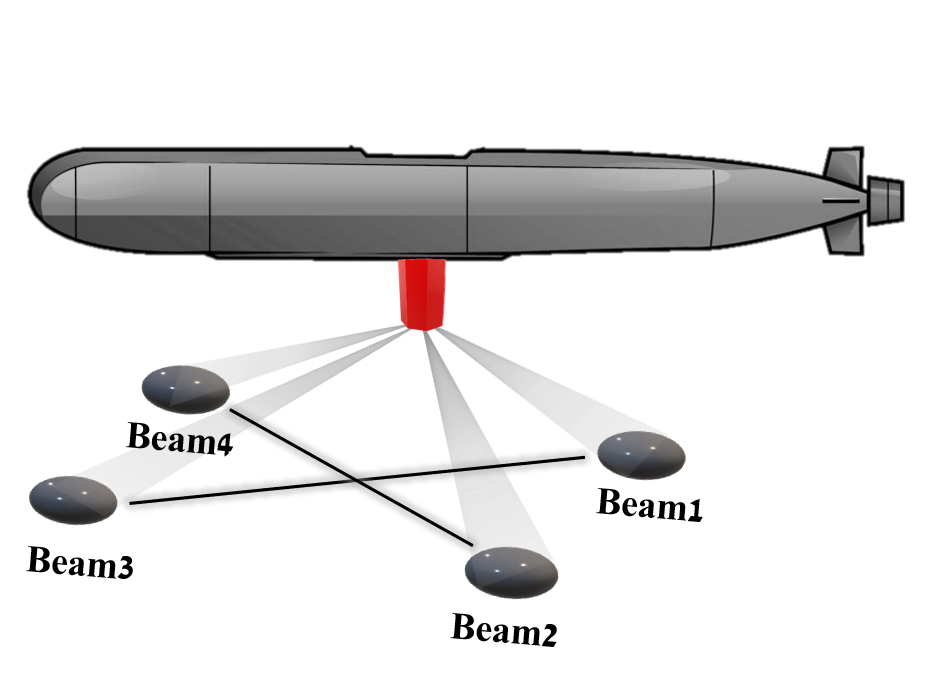}
    \caption{Illustration of a DVL mounted on an AUV, transmitting four acoustic beams to the seabed.
}
    \label{fig:DVL}
\end{figure}

\noindent
The INS provides a complete navigation solution, including the platform's position, velocity, and orientation, by performing dead-reckoning. This process involves integrating the specific force measured by the accelerometer and the angular velocity provided by the gyroscope to produce the navigation solution. However, due to inherent errors such as sensor biases and white noise, the integration is not optimal and results in error accumulation over time. The DVL operates by transmitting acoustic beams toward the seabed. As its name suggests, it utilizes the Doppler effect and measures the frequency shift of the returning signals to calculate the platform's velocity relative to the seabed. The DVL provides velocity measurements at a relatively low rate, typically around 1 Hz, compared to the INS, which operates at 100 Hz. However, DVL measurements are significantly more accurate than those provided by the INS. A nonlinear filter can be utilized to fuse data from these two sensors and achieve an improved navigation solution. In our case, we have chosen to employ the EKF. The EKF effectively integrates the high-frequency but drift-prone data from the INS with the low-frequency yet highly accurate velocity measurements from the DVL, resulting in enhanced overall navigation accuracy and robustness. 
\\ \noindent
To formulate the INS/DVL sensor fusion, we define the error-state vector as \( \delta \boldsymbol{x} \in \mathbb{R}^{12\times1} \). The error-state dynamics are described by the following equation:
\begin{equation}\label{eqn:f}
\centering
\delta \dot{\boldsymbol{x}}= \mathbf{F}\delta\boldsymbol{ x} +\mathbf{G}\boldsymbol{\omega}
\end{equation}
where \( \boldsymbol{\omega} \in \mathbb{R}^{12\times1} \) represents the system noise vector, \( \mathbf{F} \in \mathbb{R}^{12\times12} \) denotes the system matrix, and \( \mathbf{G} \in \mathbb{R}^{12\times12} \) is the system noise distribution matrix. The system is subject to multiple independent noise sources, each assumed to follow a zero-mean Gaussian distribution. The error-state vector is defined as:
\begin{equation}\label{eqn:dx}
\centering
\delta{\boldsymbol{x}}= \left[\delta{(\boldsymbol{ v^{n}}})^{T}\quad (\boldsymbol{ \epsilon^{n}})^{T}\quad \delta{\boldsymbol{ b_{a}}}^{T}\quad \delta\boldsymbol{ b_{g}}^{T}\right]^{T} \in \mathbb{R}^{12\times1}
\end{equation}
In this formulation:
\begin{itemize}
    \item \( \delta\boldsymbol{ v^{n}} \in \mathbb{R}^{3\times1} \) represents the velocity error-state in the navigation frame
    \item \( \boldsymbol{\epsilon} \in \mathbb{R}^{3\times1} \) denotes the misalignment error
    \item \( \delta\boldsymbol{b_{a}} \in \mathbb{R}^{3\times1} \) corresponds to the accelerometer bias residual error
    \item \( \delta\boldsymbol{b_{g}} \in \mathbb{R}^{3\times1} \) represents the gyroscope bias residual error
\end{itemize}
\noindent
While the INS operates and no DVL measurements are available, the EKF remains in the prediction phase, where the \textit{a priori} mean is defined as:

\begin{equation}\label{eqn:apriorizero}
\centering
\delta\boldsymbol{x}^{-}=0
\end{equation}
Next, the state covariance is propagated using the known model, which, in this case, corresponds to the nonlinear equations of motion of the INS:

\begin{equation}\label{eqn:pred}
\centering
\mathbf{P}^{-}_{k}=\mathbf{\Phi}_{k-1}\mathbf{P}^{+}_{k-1}\mathbf{\Phi}_{k-1}^{T}+\mathbf{Q}_{k-1}
\end{equation}
In this context, \( \mathbf{P}^{-}_{k} \) represents the predicted state covariance at time \( k \), while \( \mathbf{P}^{+}_{k-1} \) corresponds to the updated state covariance from the previous time step \( k-1 \). The matrix \( \mathbf{Q}_{k-1} \) represents the discrete process noise covariance, while the transition matrix \( \mathbf{\Phi}_{k-1} \) is typically obtained through a power-series expansion of the system matrix \( \mathbf{F} \) over the sampling interval \( \tau_{s} \).
\\ \noindent
Once DVL measurements are available, the EKF performs the update step, during which the state estimates are refined using the following expressions:
\begin{equation}\label{gain}
\mathbf{K}_{k} = \mathbf{P}^{-}_{k}\mathbf{H}^{T}_{k}\left(\mathbf{H}_{k}\mathbf{P}^{-}_{k}\mathbf{H}^{T}_{k} +\mathbf{R}_{k} \right)^{-1}
\end{equation}
\begin{equation}\label{postriori}
\mathbf{P}^{+}_{k} =[\mathbf{I}-\mathbf{K}_{k}\mathbf{H}_{k} ] \mathbf{P}^{-}_{k}
\end{equation}
\begin{equation}\label{dz}
\delta\boldsymbol{{x}}^{+}_{k}=\mathbf{K}_{k}\delta\boldsymbol{{z}}_{k}
\end{equation}
In these equations, \( \mathbf{K}_{k} \) denotes the Kalman gain, which determines the optimal weighting between new sensor data and predicted values from the system dynamics. The terms \( \mathbf{H}_{k} \) and \( \mathbf{R}_{k} \) refer to the observation matrix and the covariance of the measurement noise, respectively. Finally, \( \delta\boldsymbol{{x}}^{+}_{k} \) represents the corrected error-state estimate, while \( \mathbf{P}^{+}_{k} \) denotes the updated error covariance matrix. All the variables mentioned above are well-defined and can be found in the literature, for example, in \cite{farrell2008aided,Groves2013}.
\\ \noindent
To incorporate velocity observations into the EKF model, it is essential to first analyze the geometric configuration of the DVL's operation. The DVL is mounted within the AUV with its transducers oriented toward the seabed. It employs an "$\times$" pattern configuration, commonly referred to as the "Janus Doppler configuration" \cite{cohen2024kit}. An illustration can be seen in Fig.\ref{fig:DVL}. In this configuration, the transducers are positioned at specific yaw and pitch angles relative to the DVL sensor's frame, which typically differs from the platform’s body frame. The relationship is given by:
\begin{equation}\label{eqn:4}
    \centering
        \boldsymbol{b}_{\dot{\imath}}=
        \begin{bmatrix} 
        \cos{\psi_{\dot{\imath}}}\sin{\theta}\quad
        \sin{\psi_{\dot{\imath}}}\sin{\theta}\quad
        \cos{\theta}
    \end{bmatrix}_{1\times3}
\end{equation} 
where \( \boldsymbol{b}_{\dot{\imath}} \) (for \( \dot{\imath}=1,2,3,4 \)) corresponds to the beam number, and \( \psi \) and \( \theta \) represent the yaw and pitch angles relative to the body frame. The pitch angle is predetermined by the manufacturer and remains constant across all beams. The yaw angle, on the other hand, can be defined as follows:
\begin{equation}\label{eqn:5}
    \centering
        \psi_{\dot{\imath}}=(\dot{\imath}-1)\cdot90^{\circ}+45^{\circ}\;,\; \dot{\imath}=1,2,3,4
\end{equation}
Let \( \boldsymbol{v}^{DVL} \) denote the velocity vector in the DVL frame, and let \( \mathbf{T} \) be the transformation matrix that maps it to the beam-aligned velocity vector \( \boldsymbol{v}^{Beam} \). This relationship is expressed as:
\begin{equation}\label{eqn:6}
    \centering
        \boldsymbol{v}^{Beam}=\mathbf{T}\boldsymbol{v}^{DVL} ,\quad
        \mathbf{T}=
        \begin{bmatrix} \boldsymbol{b}_{1}\\\boldsymbol{b}_{2}\\\boldsymbol{b}_{3}\\\boldsymbol{b}_{4}\\
    \end{bmatrix}_{4\times3}
\end{equation} 
The beam measurements are affected by various errors, which can be modeled as:
\begin{equation}\label{eqn:7}
    \centering
        \tilde{\boldsymbol{v}}^{Beam}= \mathbf{T}[\boldsymbol{v}^{DVL}(\boldsymbol{1}+\boldsymbol{s}_{DVL})]+\boldsymbol{b}_{DVL}+\boldsymbol{n}_{DVL}
\end{equation}
where, \( \boldsymbol{b}_{DVL} \in \mathbb{R}^{4\times1} \) represents the bias vector, \( \boldsymbol{s}_{DVL} \in \mathbb{R}^{3\times1} \) denotes the scale factor vector, and \( \boldsymbol{n}_{DVL} \in \mathbb{R}^{4\times1} \) is modeled as zero-mean Gaussian noise. Once the raw measurements are obtained, the velocity \( \boldsymbol{v}^{DVL} \) can be estimated by minimizing the following cost function:
\begin{equation}\label{eqn:8}
    \centering
        \hat{\boldsymbol{v}}^{DVL}=
        \underset{\boldsymbol{v}^{DVL}}{\argmin}{\mid\mid\tilde{\boldsymbol{v}}^{Beam}-\mathbf{T}\boldsymbol{v}^{DVL} \mid\mid}^{2}.
\end{equation} 
The solution to this least-squares problem is computed by multiplying the measurements by the pseudo-inverse of the transformation matrix \( \mathbf{T} \) \cite{bar2004estimation}:
\begin{equation}\label{eqn:99}
    \centering
        \hat{\boldsymbol{v}}^{DVL}=(\mathbf{T}^{T}\mathbf{T})^{-1}\mathbf{T}^{T}\tilde{\boldsymbol{v}}^{Beam}.
\end{equation}
This outcome can be integrated into the EKF in a loosely coupled manner \cite{Groves2013}.
\section{Multi-Output Gaussian Process Regression}\label{sec:MOGPR}
\noindent
Gaussian Process (GP) regression is a powerful non-parametric Bayesian approach for modeling functions. Given a training set defined as:
\begin{equation}
\centering
 \mathcal{D} = \{ (\mathbf{x}_i, y_i) \mid i = 1, \dots, n \}   
\end{equation}
\noindent where:
\begin{itemize}
    \item \( \mathbf{x}_i \in \mathbb{R}^D \) represents the input vector
    \item \( y_i \in \mathbb{R} \) is the corresponding output (target)
    \item \( n \) denotes the number of observations
\end{itemize}
\noindent The entire training set can be represented in matrix form as:
\begin{equation}
\centering
   \mathcal{D} = (\mathbf{X}, \mathbf{y}),
\end{equation}
\noindent where the design matrix \( \mathbf{X} \) and the target vector \( \mathbf{y} \) are defined as:
\begin{equation}
\centering
\mathbf{X} =
\begin{bmatrix}
    x_{11} & x_{12} & \dots  & x_{1D} \\
    x_{21} & x_{22} & \dots  & x_{2D} \\
    \vdots & \vdots & \ddots & \vdots \\
    x_{n1} & x_{n2} & \dots  & x_{nD}
\end{bmatrix}, \quad \mathbf{y} =
\begin{bmatrix}
    y_1 \\
    y_2 \\
    \vdots \\
    y_n
\end{bmatrix}
\end{equation}

\noindent The objective in regression tasks is to infer the relationship between inputs and outputs by modeling the conditional distribution \( p(\mathbf{y} \mid \mathbf{X}) \), without explicitly modeling the input distribution.

\noindent A GP is fully defined by its mean function \( m(\mathbf{x}) \) and covariance function \( c(\mathbf{x}, \mathbf{x'}) \), which together define the distribution over possible functions. Formally, for a real-valued function \( f(\mathbf{x}) \), the mean and covariance functions are given by:
\begin{equation}
    m(\mathbf{x}) = \mathbb{E}[f(\mathbf{x})],
\end{equation}
\begin{equation}
    c(\mathbf{x}, \mathbf{x'}) = \mathbb{E}[(f(\mathbf{x}) - m(\mathbf{x}))(f(\mathbf{x'}) - m(\mathbf{x'}))].
\end{equation}
The Gaussian process is then expressed as:
\begin{equation}
    f(\mathbf{x}) \sim \mathcal{GP}(m(\mathbf{x}), c(\mathbf{x}, \mathbf{x'})).
\end{equation}
For simplicity, the mean function is often assumed to be zero.

\noindent In machine learning applications, the random variables in a GP represent function values at different input locations. Unlike conventional parametric models, a GP defines a distribution over functions without requiring an explicit functional form. Instead, it captures correlations between function values via the covariance function \( c(\mathbf{x}, \mathbf{x'}) \).

\noindent In practice, function values are not directly observable and are subject to added stochastic noise:
\begin{equation}
    y = f(\mathbf{x}) + \varepsilon,
\end{equation}
where \( \varepsilon \sim \mathcal{N}(0, \sigma_n^2) \) represents Gaussian observation noise.

\noindent Given the training data \(
\mathcal{D} = (\mathbf{X}, \mathbf{y}),
\) and test locations \( \mathbf{X}_* \), the joint distribution of observed and predicted values is:
\begin{equation}
    \begin{bmatrix} 
        \mathbf{y} \\ 
        f_* 
    \end{bmatrix} 
    \sim \mathcal{N} \left(
    \mathbf{0}, 
    \begin{bmatrix}
        C(\mathbf{X}, \mathbf{X}) + \sigma_n^2 I & C(\mathbf{X}, \mathbf{X}_*) \\
        C(\mathbf{X}_*, \mathbf{X}) & C(\mathbf{X}_*, \mathbf{X}_*)
    \end{bmatrix}
    \right).
\end{equation}
where the predictive distribution \( f_* \triangleq f(\mathbf{x}_*) \) at \( \mathbf{x}_* \).
Using standard Gaussian conditioning rules, the predictive distribution for test points is given by:
\begin{equation}\label{mean_gpr}
    \mathbb{E}[f_* | \mathbf{X}, \mathbf{y}] = C(\mathbf{X}_*, \mathbf{X}) [C(\mathbf{X}, \mathbf{X}) + \sigma_n^2 I]^{-1} \mathbf{y},
\end{equation}
\begin{equation}\label{cov_gpr}
    \begin{aligned}
        \text{cov}(f_*) &= C(\mathbf{X}_*, \mathbf{X}_*) \\
        &\quad - C(\mathbf{X}_*, \mathbf{X}) [C(\mathbf{X}, \mathbf{X}) + \sigma_n^2 I]^{-1} C(\mathbf{X}, \mathbf{X}_*).
    \end{aligned}
\end{equation}
If there are \( n \) training points and \( n_* \) test points, the kernel matrix \( C(X, X_*) \) is an \( n \times n_* \) matrix that contains the covariances computed for all pairs of training and test points using the kernel function \( c(\cdot, \cdot) \). The same notation applies to other kernel matrices \( C(X, X) \), \( C(X_*, X_*) \), and \( C(X_*, X) \), where \( C \) represents the kernel matrix and \( c \) denotes the underlying kernel function. Equations \eqref{mean_gpr}-\eqref{cov_gpr} form the foundation of Gaussian process regression, enabling robust function approximation and uncertainty quantification \cite{williams2006gaussian,murphy2023probabilistic}.

\noindent Extending single-output GPR to MOGPR aims to model multiple correlated outputs simultaneously while preserving dependencies across different input and output dimensions. In order to achieve this, separable kernels are used to decouple the contributions of the input and output spaces. The kernel matrix for a dataset \( \mathbf{X} \) can then be written as:
\begin{equation}\label{Kronecker}
    \mathbf{C}(\mathbf{X}, \mathbf{X}) =  \mathbf{B} \otimes c(\mathbf{X}, \mathbf{X})
\end{equation}
where \( \otimes \) denotes the Kronecker product between matrices. This formulation enhances computational efficiency by exploiting the structure of the covariance matrix. The simplest case occurs when \( \mathbf{B} \) is set as an identity matrix \( \mathbf{I} \), implying an assumption of independent outputs \cite{alvarez2012kernels}.

\section{Proposed Approach}\label{sec:sa}
\noindent
Our proposed approach aims to provide a data-driven alternative to the LS solution, which is commonly used in the literature and industry, for estimating velocity vector and its associated variance. The motivation stems from the fact that DVL measurements are, in practice, subject to bias, even after careful calibration. This bias renders the LS solution suboptimal and inefficient. This inherent bias creates an opportunity for improvement through a more robust modeling approach.
\\
\noindent To apply the MOGPR framework to DVL velocity estimation, we construct the training set as follows:
\begin{equation}
\centering
 \mathcal{D}^{DVL} = \{ (\tilde{\boldsymbol{v}}^{Beam}_i, \boldsymbol{v}^{DVL}_i) \mid i = 1, \dots, n \},   
\end{equation}
where \( \boldsymbol{v}^{DVL}_i \) and \( \tilde{\boldsymbol{v}}^{Beam}_i \) are defined in \eqref{eqn:6} and \eqref{eqn:7}, respectively. The training set pairs beam velocity measurements \( \tilde{\boldsymbol{v}}^{Beam}_i \) with their corresponding DVL velocities \( \boldsymbol{v}^{DVL}_i \), enabling the model to learn the mapping between these variables.
\noindent To enhance the flexibility of the MOGPR model, we employ a combination of three automatic relevance determination (ARD) kernels by adding them:

\noindent The ARD squared exponential kernel:
\begin{equation}
    c(x_i, x_j | \boldsymbol{\theta}) = \sigma_f^2 \exp \left( -\frac{1}{2} \sum_{m=1}^{4} \frac{(x_{im} - x_{jm})^2}{\sigma_m^2} \right)
\end{equation}

\noindent The ARD Matérn \( \frac{3}{2} \) kernel:
\begin{equation}
    \begin{aligned}
        c(x_i, x_j | \boldsymbol{\theta}) &= \sigma_f^2 \left( 1 + \sqrt{3} \, r \right) \exp \left( -\sqrt{3} \, r \right),
        \\
        r &= \sqrt{\sum_{m=1}^{4} \frac{(x_{im} - x_{jm})^2}{\sigma_m^2}}
    \end{aligned}
\end{equation}

\noindent with the ARD rational quadratic kernel defined by:
\begin{equation}
    c(x_i, x_j | \boldsymbol{\theta}) = \sigma_f^2 \left( 1 + \frac{1}{2} \sum_{m=1}^{4} \frac{(x_{im} - x_{jm})^2}{\sigma_m^2} \right)^{-1}
\end{equation}

\noindent ARD kernels allow the model to automatically determine the relevance of each input feature by assigning individual length scales \( \sigma_m \) to each input dimension. This capability makes ARD kernels particularly effective for feature selection and dimensionality reduction, enabling the model to focus on the most informative features while down-weighting less relevant ones. Before being combined, each kernel is adjusted to the MOGPR framework according to \eqref{Kronecker}, with \( \mathbf{B} \) set as the identity matrix, as the outputs represent uncorrelated velocities.
\begin{figure}[h!]
    \centering
    \begin{subfigure}{\columnwidth}
        \centering
        \includegraphics[width=\linewidth]{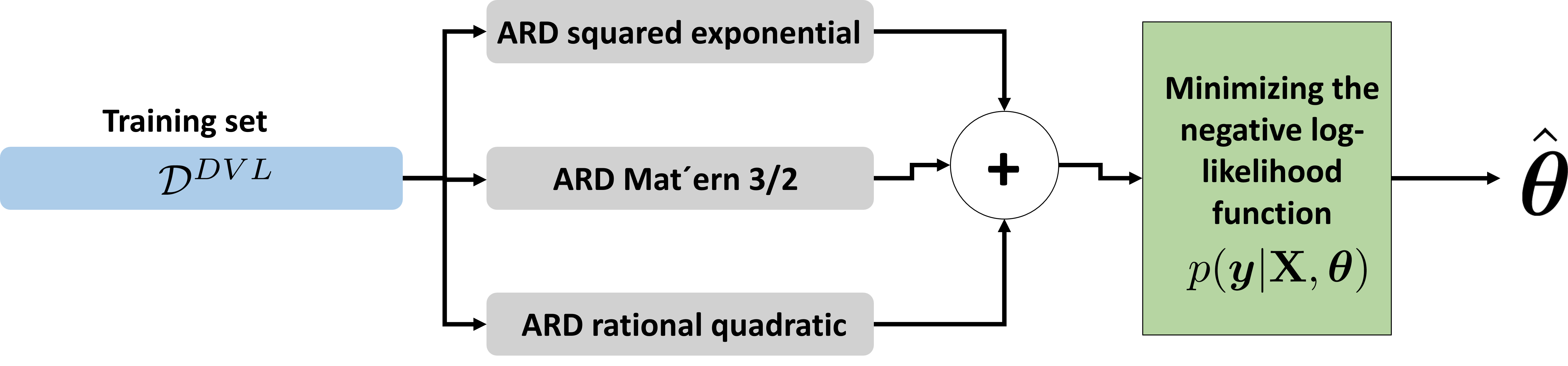}
        \caption{}
        \label{fig:sub1}
    \end{subfigure}
    
    \vspace{0.5cm} 
    
    \begin{subfigure}{\columnwidth}
        \centering
        \includegraphics[width=\linewidth]{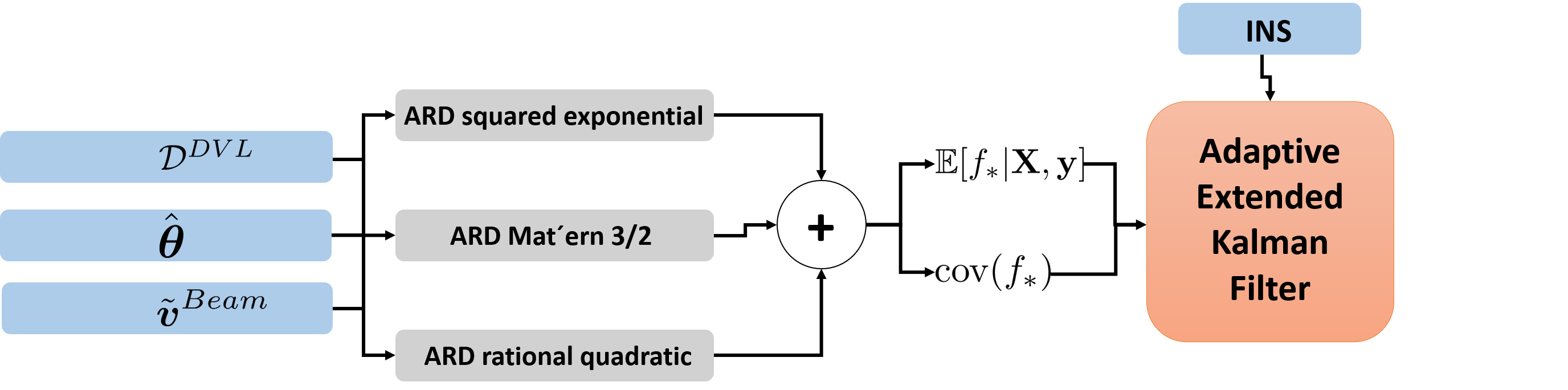}
        \caption{}
        \label{fig:sub2}
    \end{subfigure}
    
    \caption{The block diagram of the proposed approach begins with (a), where the training data is fed into each of the ARD kernels, which are then combined. The hyperparameters within the kernels are optimized by minimizing the negative log-likelihood. Once the optimized parameter vector is obtained, the process moves to (b), where a test point is incorporated to generate the velocity prediction along with its associated covariance. These values are then used within the EKF for state estimation.
}
    \label{fig.block}
\end{figure}
\\ \noindent
Next, we construct a learnable variables vector \( \boldsymbol{\theta} = \{\sigma_f, \dots, \sigma_m, \dots, \sigma_n\} \), which can be optimized by minimizing the negative log-likelihood function \( p(\boldsymbol{y} | \mathbf{X}, \boldsymbol{\theta}) \). Further details on this optimization process can be found in \cite{williams2006gaussian}. In this work, the optimization was performed using gradient descent and the ADAM optimizer, with a learning rate of 0.1 and first and second moment coefficients \( \beta_1 = 0.9 \) and \( \beta_2 = 0.999 \), respectively, over 50 iterations \cite{kingma2014adam}.
\\ \noindent
Once the above steps are completed, the MOGPR solution, defined by \eqref{mean_gpr} and \eqref{cov_gpr}, is obtained and can be utilized in a loosely coupled approach within the error-state EKF. Moreover, the MOGPR provides uncertainty in the measurements, which can be used as an adaptive measurement noise matrix \( \mathbf{R}_k \), further enhancing the performance of the EKF. A block diagram of the suggested approach is shown in Fig.~\ref{fig.block}.
\section{Experimental Results}\label{sec:res}
\subsection{AUV Sea Trials}
\noindent
A sea experiment was conducted in the Mediterranean Sea using an AUV to collect the data. Specifically, the "Snapir" AUV, a modified ECA Group A18D mid-size platform, was utilized. This AUV can autonomously execute missions at depths of up to three kilometers with an endurance of 21 hours.
\begin{figure}[h!]
    \centering
    \begin{subfigure}{\columnwidth}
        \centering
        \includegraphics[width=\columnwidth]{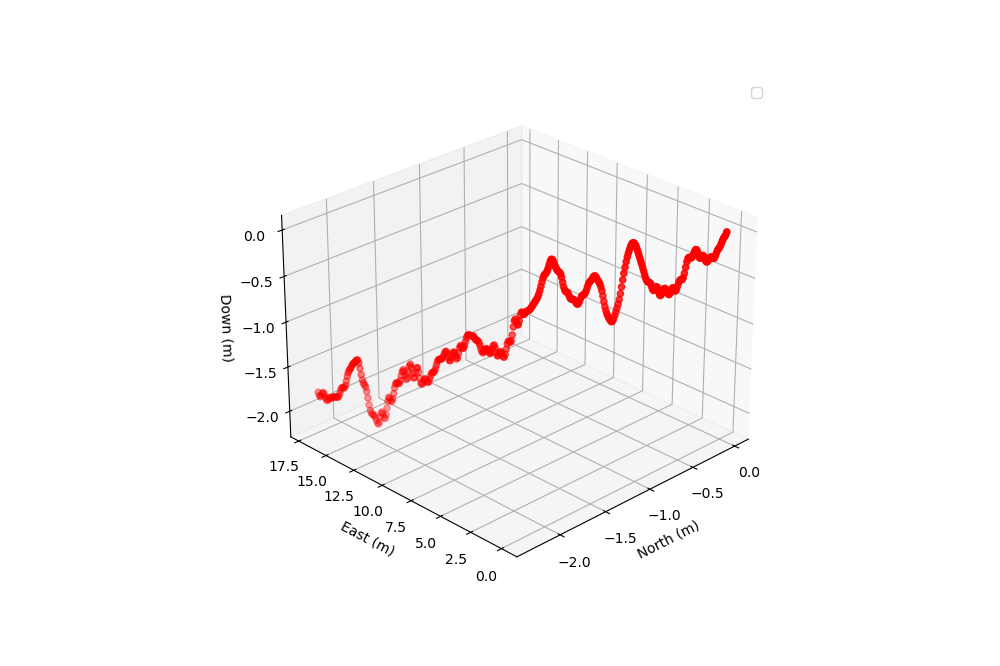}
        \caption{}
        \label{fig:sub1}
    \end{subfigure}

    \begin{subfigure}{\columnwidth}
        \centering
        \includegraphics[width=\columnwidth]{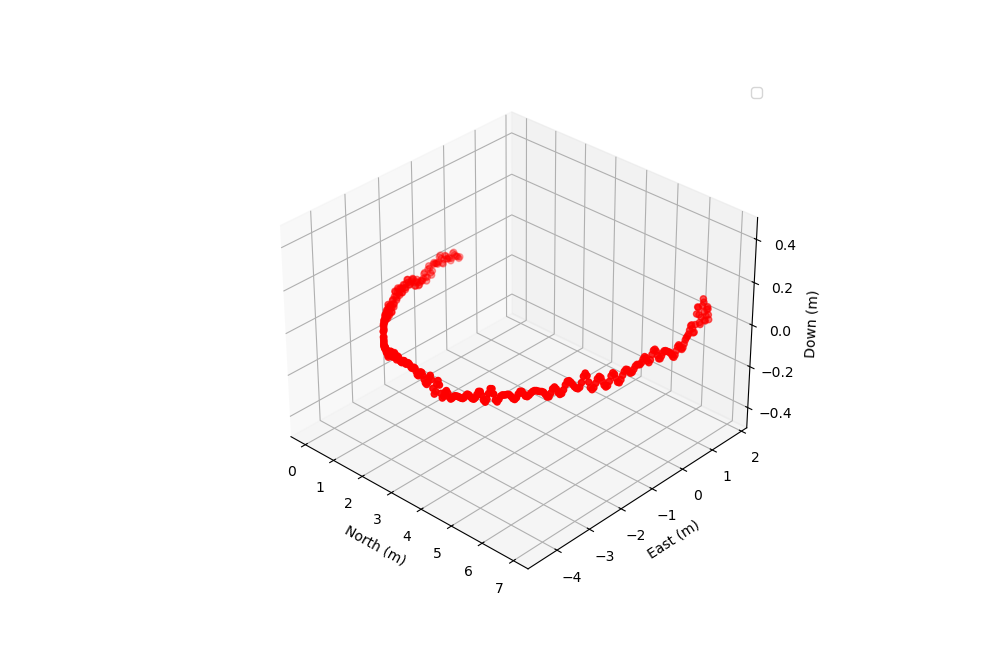}
        \caption{}
        \label{fig:sub2}
    \end{subfigure}
    
    \caption{ The two, out of the thirteen trajectories taken from \cite{cohen2024kit}, that were used for testing are trajectory number 12 (a) and trajectory number 13 (b).}
    \label{fig:traj}
\end{figure}
\noindent
The inertial sensor employed in the AUV is the iXblue Phins Subsea INS \cite{iXblue_PHINS}, a navigation-grade INS, while the DVL used is the Teledyne RDI Workhorse Navigator \cite{TeledyneMarine_DVL}. Over seven hours of data were collected, from which 13 distinct mission scenarios were carefully selected for training and testing the model. These missions are characterized by varying maneuvers, speeds, depths, and other factors, providing a robust representation of typical AUV tasks. Further details regarding the dataset are provided in \cite{cohen2024kit}. Trajectories 12 and 13 are used to test and evaluate the suggested approach and can be seen in Fig.\ref{fig:traj}.

\section{Performance}
\begin{figure}[t!]
    \centering
    \begin{subfigure}{0.8\columnwidth}
        \centering
        \includegraphics[width=\columnwidth]{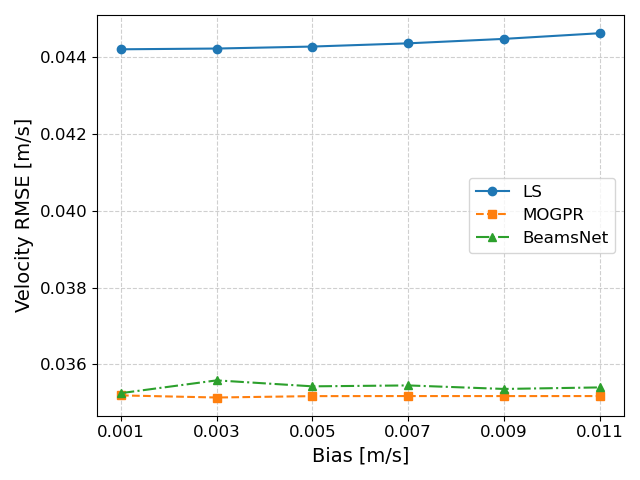}
        \caption{}
        \label{fig:sub1}
    \end{subfigure}

    \begin{subfigure}{0.8\columnwidth}
        \centering
        \includegraphics[width=\columnwidth]{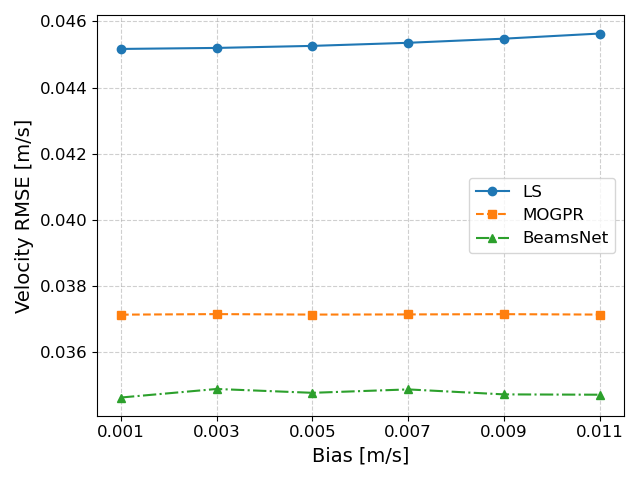}
        \caption{}
        \label{fig:sub2}
    \end{subfigure}
    
    \caption{The velocity RMSE for each of the compared approaches as a function of the added bias value is presented. In (a), the results for trajectory 12 are shown, while in (b), the corresponding results for trajectory 13 are displayed.}
    \label{fig.rmse}
\end{figure}

\noindent
The suggested approach was trained using eleven out of the thirteen trajectories. In addition to the proposed method, we compared its performance against the LS estimator and another data-driven approach named BeamsNet. BeamsNet is a deep learning method based on a one-dimensional convolutional neural network, which has been shown to provide better estimations than the LS solution. Similar to the proposed approach, BeamsNet takes the beam velocity measurements as input but also includes past DVL measurements. Further details about BeamsNet, including its architecture, hyperparameters, and implementation, can be found in \cite{cohen2022beamsnet}.
\\ \noindent
We used the error model presented in \eqref{eqn:6} to create scenarios that mimic inadequate calibration. A range of bias vectors was added, starting from 0.001~[m/s] and increasing in increments of 0.002~[m/s] up to 0.011~[m/s]. Additionally, white noise with a standard deviation of 0.02~[m/s] was included under a zero scale factor assumption. 
\noindent
First, the three approaches' root mean squared error (RMSE) was calculated relative to the ground truth data, and as a function of the added bias, the results are summarized in Fig.~\ref{fig.rmse}.
\noindent
The results demonstrate that the LS estimation error increases as the bias grows, whereas both data-driven approaches maintain a relatively constant estimation accuracy. For trajectory 12, with a bias of 0.011~[m/s], MOGPR and BeamsNet performed similarly, achieving an approximately 20\% improvement over the LS estimation. In trajectory 13, BeamsNet outperformed MOGPR by a small margin while still providing a comparable improvement relative to the LS method. A key advantage of MOGPR is that it also provides the covariance, which can be dynamically incorporated into the EKF as adaptive measurement noise in addition to accurate velocity estimation.
\\ \noindent
Next, the velocity estimation values from each approach were integrated into the error-state EKF as measurement updates. For the proposed MOGPR-based approach, the estimated covariance was also incorporated as adaptive measurement noise covariance.
\begin{figure}[b!]
    \centering
    \begin{subfigure}{\columnwidth}
        \centering
        \includegraphics[width=\linewidth]{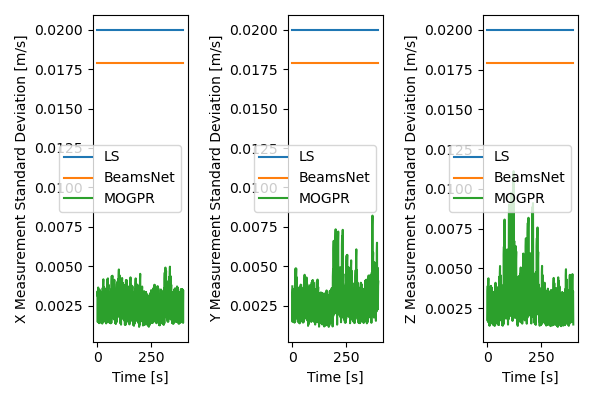}
        \caption{}
        \label{fig:sub1}
    \end{subfigure}

    \begin{subfigure}{\columnwidth}
        \centering
        \includegraphics[width=\linewidth]{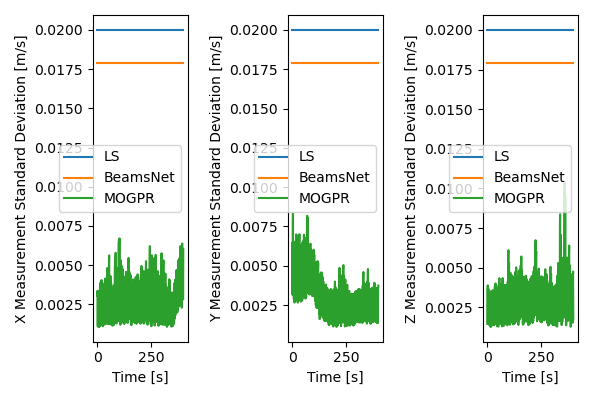}
        \caption{}
        \label{fig:sub2}
    \end{subfigure}
    
    \caption{The standard deviation values for each of the compared approaches are presented as a function of time. For LS, it is 0.02~[m/s], as specified by the sensor manufacturer. For BeamsNet, the value was determined through trial and error, while for MOGPR, it was derived from its equations. In (a), the results for trajectory 12 are presented across all axes, and in (b), the corresponding results for trajectory 13 are shown.
}
    \label{fig:mes_cov}
\end{figure}
 \noindent
The standard deviation of each method’s measurement noise matrix \( \mathbf{R}_k \) is shown in Fig.~\ref{fig:mes_cov}, illustrating how MOGPR can adaptively adjust its prediction uncertainty, whereas, in the other approaches, it remains constant and predefined. The reason the measurement noise in the LS approach remains constant is due to the linear nature of the task and the constant transformation matrix in \eqref{eqn:6}. For the BeamsNet approach, the uncertainty decreases, resulting in a constant but smaller noise value. The MOGPR also provides the smallest standard deviation over time.
\\ \noindent
Six states were observed for comparison: velocity in the north, east, and down directions within the navigation frame, as well as roll, pitch, and yaw angles transforming from the body frame to the navigation frame. The RMSE of each state is presented in Table~\ref{tab:res12} for trajectory 12 and Table~\ref{tab:res13} for trajectory 13. The results show that both data-driven approaches, when integrated into the EKF in a loosely coupled manner, outperformed the LS method across all states. Additionally, when comparing the norm of the RMSE vector across methods, for trajectory 12, the MOGPR approach outperformed the LS by at least 25\% in terms of velocity RMSE and at least 42\% in terms of angle RMSE. When compared to the second data-driven approach, BeamsNet, MOGPR achieved at least a 15.8\% improvement in velocity RMSE and at least a 45\% improvement in angle RMSE. The larger improvement in angle RMSE is primarily due to the yaw error, while BeamsNet slightly outperformed MOGPR in roll and pitch estimation. For trajectory 13, the data-driven approaches once again outperformed the model-based LS approach across all states. MOGPR showed at least a 13\% improvement over LS in velocity RMSE and at least an 18\% improvement in angle RMSE. When comparing MOGPR to BeamsNet, BeamsNet slightly outperformed MOGPR in both velocity and angle RMSE for this trajectory.
\begin{table}[h!]
\centering
\caption{The RMSE values for each velocity and orientation angle state as a function of the added bias are presented for each of the loosely coupled EKF approaches, namely LS, BeamsNet, and MOGPR, for trajectory 12.}
\label{tab:res12}
\resizebox{\columnwidth}{!}{%
\begin{tabular}{|cc|c|c|c|c|c|c|}
\hline
\multicolumn{2}{|c|}{Bias Value {[}m/s{]}} &
  \begin{tabular}[c]{@{}c@{}}Velocity\\ North\\ {[}m/s{]}\end{tabular} &
  \begin{tabular}[c]{@{}c@{}}Velocity\\ East\\ {[}m/s{]}\end{tabular} &
  \begin{tabular}[c]{@{}c@{}}Velocity\\ Down\\ {[}m/s{]}\end{tabular} &
  Roll {[}$^\circ${]} &
  Pitch {[}$^\circ${]} &
  Yaw {[}$^\circ${]} \\ \hline
      & LS       & 0.076          & 0.054          & 0.028          & 0.591          & 0.170          & 1.461          \\
0.001 & BeamsNet & 0.070          & 0.046          & 0.023          & \textbf{0.211} & \textbf{0.154} & 1.974          \\
      & MOGPR    & \textbf{0.057} & \textbf{0.037} & \textbf{0.023} & 0.409          & 0.159          & \textbf{0.714} \\ \hline
      & LS       & 0.076          & 0.054          & 0.030          & 0.593          & 0.169          & 1.461          \\
0.003 & BeamsNet & 0.070          & 0.046          & \textbf{0.022} & \textbf{0.199} & \textbf{0.157} & 1.966          \\
      & MOGPR    & \textbf{0.057} & \textbf{0.037} & 0.024          & 0.413          & 0.159          & \textbf{0.716} \\ \hline
      & LS       & 0.076          & 0.054          & 0.031          & 0.595          & 0.169          & 1.462          \\
0.005 & BeamsNet & 0.073          & 0.046          & \textbf{0.022} & \textbf{0.196} & \textbf{0.155} & 2.023          \\
      & MOGPR    & \textbf{0.057} & \textbf{0.037} & 0.023          & 0.410          & 0.159          & \textbf{0.716} \\ \hline
      & LS       & 0.076          & 0.054          & 0.032          & 0.597          & 0.169          & 1.463          \\
0.007 & BeamsNet & 0.071          & 0.046          & \textbf{0.023} & \textbf{0.215} & \textbf{0.154} & 1.992          \\
      & MOGPR    & \textbf{0.057} & \textbf{0.037} & 0.024          & 0.412          & 0.159          & \textbf{0.718} \\ \hline
      & LS       & 0.076          & 0.054          & 0.033          & 0.600          & 0.169          & 1.463          \\
0.009 & BeamsNet & 0.073          & 0.046          & \textbf{0.022} & \textbf{0.198} & \textbf{0.154} & 2.031          \\
      & MOGPR    & \textbf{0.057} & \textbf{0.037} & 0.024          & 0.411          & 0.159          & \textbf{0.716} \\ \hline
      & LS       & 0.076          & 0.054          & 0.035          & 0.602          & 0.169          & 1.464          \\
0.011 & BeamsNet & 0.073          & 0.046          & \textbf{0.022} & \textbf{0.194} & \textbf{0.155} & 2.017          \\
      & MOGPR    & \textbf{0.057} & \textbf{0.037} & 0.023          & 0.412          & 0.159          & \textbf{0.717} \\ \hline
\end{tabular}%
}
\end{table}
\begin{table}[h!]
\centering
\caption{The RMSE values for each velocity and orientation angle state as a function of the added bias are presented for each of the loosely coupled EKF approaches, namely LS, BeamsNet, and MOGPR, for trajectory 13.
}
\label{tab:res13}
\resizebox{\columnwidth}{!}{%
\begin{tabular}{|cc|c|c|c|c|c|c|}
\hline
\multicolumn{2}{|c|}{Bias Value {[}m/s{]}} &
  \begin{tabular}[c]{@{}c@{}}Velocity\\ North\\ {[}m/s{]}\end{tabular} &
  \begin{tabular}[c]{@{}c@{}}Velocity\\ East\\ {[}m/s{]}\end{tabular} &
  \begin{tabular}[c]{@{}c@{}}Velocity\\ Down\\ {[}m/s{]}\end{tabular} &
  Roll {[}$^\circ${]} &
  Pitch {[}$^\circ${]} &
  Yaw {[}$^\circ${]} \\ \hline
      & LS       & 0.082          & 0.07           & 0.023          & 0.107          & 0.118          & 2.397          \\
0.001 & BeamsNet & \textbf{0.064} & 0.058          & 0.023          & 0.142          & 0.116          & \textbf{1.632} \\
      & MOGPR    & 0.074          & \textbf{0.057} & \textbf{0.020} & \textbf{0.101} & \textbf{0.110} & 1.929          \\ \hline
      & LS       & 0.082          & 0.07           & 0.024          & 0.106          & 0.118          & 2.395          \\
0.003 & BeamsNet & \textbf{0.061} & \textbf{0.055} & 0.023          & 0.138          & 0.115          & \textbf{1.442} \\
      & MOGPR    & 0.074          & 0.057          & \textbf{0.020} & \textbf{0.101} & \textbf{0.110} & 1.926          \\ \hline
      & LS       & 0.082          & 0.07           & 0.024          & 0.106          & 0.118          & 2.393          \\
0.005 & BeamsNet & \textbf{0.064} & 0.059          & 0.023          & 0.142          & 0.116          & \textbf{1.637} \\
      & MOGPR    & 0.074          & \textbf{0.057} & \textbf{0.020} & \textbf{0.101} & \textbf{0.110} & 1.930          \\ \hline
      & LS       & 0.082          & 0.070          & 0.025          & 0.106          & 0.118          & 2.392          \\
0.007 & BeamsNet & \textbf{0.065} & 0.059          & 0.023          & 0.142          & 0.117          & \textbf{1.661} \\
      & MOGPR    & 0.074          & \textbf{0.057} & \textbf{0.020} & \textbf{0.101} & \textbf{0.110} & 1.931          \\ \hline
      & LS       & 0.082          & 0.07           & 0.025          & 0.105          & 0.118          & 2.39           \\
0.009 & BeamsNet & \textbf{0.064} & 0.059          & 0.023          & 0.141          & 0.116          & \textbf{1.628} \\
      & MOGPR    & 0.074          & \textbf{0.057} & \textbf{0.021} & \textbf{0.101} & \textbf{0.110} & 1.927          \\ \hline
      & LS       & 0.082          & 0.07           & 0.026          & 0.105          & 0.118          & 2.388          \\
0.011 & BeamsNet & \textbf{0.064} & 0.059          & 0.023          & 0.141          & 0.116          & \textbf{1.622} \\
      & MOGPR    & 0.074          & \textbf{0.057} & \textbf{0.020} & \textbf{0.101} & \textbf{0.110} & 1.933          \\ \hline
\end{tabular}%
}
\end{table}

\section{Conclusions}\label{sec:con}
\noindent
AUVs are typically required to conduct long-duration missions at depths beyond human reach. Although these platforms are equipped with high-end sensors, they must be carefully calibrated to achieve optimal performance. However, such calibrations are often challenging to perform and, in some cases, are neglected. Even with meticulous calibration, sensors are never entirely bias-free. When this occurs, the model-based LS estimation is no longer, by definition, an optimal estimator, leaving room for improvement. In the case of DVL velocity estimation, the previously proposed data-driven model, BeamsNet, has demonstrated the potential to achieve such improvements.
\\ \noindent
In this work, we leveraged the capabilities of GPR and, through a MOGPR, estimated the AUV velocity using raw DVL measurements as an alternative to the LS approach. The MOGPR not only provides velocity estimates but also yields an adaptive covariance that reflects the model’s confidence in its predictions. This covariance was incorporated as the measurement noise covariance matrix in the update phase of the error-state EKF, enabling an adaptive EKF framework for a complete AUV navigation solution. The results show that, beyond improving the velocity estimation by approximately 20\%, MOGPR also enhances the overall navigation solution within the EKF, leading to improvements in other states, such as orientation angles. Additionally, MOGPR outperformed BeamsNet in most state estimations.
\\ \noindent
While data-driven approaches hold promise for real-time systems requiring reliability, their limitations are well known. MOGPR requires storing the entire training set and involves inverting a high-dimensional matrix, which is computationally expensive. These challenges and the necessity of acquiring sufficiently generalized and high-quality training data must be addressed before these methods can be fully integrated into real-time navigation systems.

\section*{Acknowledgment}
\noindent
N.C. is supported by the Maurice Hatter Foundation and the University of Haifa presidential scholarship for outstanding students on a direct Ph.D. track.
\bibliographystyle{IEEEtran}
\bibliography{bio.bib}
\end{document}